\documentclass[conference]{IEEEtran}
\IEEEoverridecommandlockouts
\pdfoutput=1
\usepackage{cite}
\usepackage{amsmath,amssymb,amsfonts}
\usepackage{algorithmic}
\usepackage{graphicx}
\usepackage{textcomp}
\usepackage{xcolor}
\def\BibTeX{{\rm B\kern-.05em{\sc i\kern-.025em b}\kern-.08em
    T\kern-.1667em\lower.7ex\hbox{E}\kern-.125emX}}
\begin{document}

\title{VRChain: A Blockchain-Enabled Framework for Visual Homing and Navigation Robots 
}

\author{

\IEEEauthorblockN{Mohamed Rahouti}
\IEEEauthorblockA{\textit{CIS Dept.} \\
\textit{Fordham Univ.}\\
NY, USA \\
mrahouti@fordham.edu}
\and
\IEEEauthorblockN{Damian Lyons}
\IEEEauthorblockA{\textit{CIS Dept.} \\
\textit{Fordham Univ.}\\
NY, USA \\
dlyons@fordham.edu}
\and
\IEEEauthorblockN{Lesther Santana}
\IEEEauthorblockA{\textit{CIS Dept.} \\
\textit{Fordham Univ.}\\
NY, USA \\
lsantanacarmona@fordham.edu}

}

\maketitle

\begin{abstract}
Visual homing is a lightweight approach to robot visual navigation. Based upon stored visual information of a home location, the navigation back to this location can be accomplished from any other location in which this location is visible by comparing home to the current image. However, a key challenge of visual homing is that the target home location must be within the robot's field of view (FOV) to start homing. Therefore, this work addresses such a challenge by integrating blockchain technology into the visual homing navigation system. Based on the decentralized feature of blockchain, the proposed solution enables visual homing robots to share their visual homing information and synchronously access the stored data (visual homing information) in the decentralized ledger to establish the navigation path. The navigation path represents a per-robot sequence of views stored in the ledger. If the home location is not in the FOV, the proposed solution permits a robot to find another robot that can see the home location and travel towards that desired location. The evaluation results demonstrate the efficiency of the proposed framework in terms of end-to-end latency, throughput, and scalability.
\end{abstract}


\section{Introduction} \label{sec:introduction}

Visual homing is a lightweight approach to visual navigation for a mobile robot. Using stored visual information of a target 'home' location, visual homing can be used to navigate back to this home location from any other location in which the home location is visible by comparing the home image to the current image \cite{gaussier2000visual}. It does not require a stored map of the environment and can be combined with an obstacle avoidance functionality for generality. This makes visual homing very attractive for robot platforms with a low computational capacity such as small UAV drones and ground robots \cite{castello2018blockchain}. It is also attractive for applications in which global map information may not be available, e.g., GPS-denied environments or rapidly changing environments.  A robot might store multiple different home location images related to tasks or activities it needs to perform, or the images may be transmitted to it from another robot or camera system.

A key limitation of visual homing is that the home location must be within the field of view (FOV) of the robot to start homing, and this restricts the use of the method to just the locality of the home location. One solution to address such a limitation is by breaking a path into ``line of sight" segments, or using a topological stored map with the edges labeled with intermediate home images. However, this additional map requirement may be problematic for lightweight implementations or GPS-denied or rapidly changing environments. In this paper, we propose to address the issue of the home location not being in the FOV by considering the case where the robot is part of a team of robots and by leveraging camera information from other robots or camera resources in the team.

\begin{figure}[htbp]
\centerline{\includegraphics{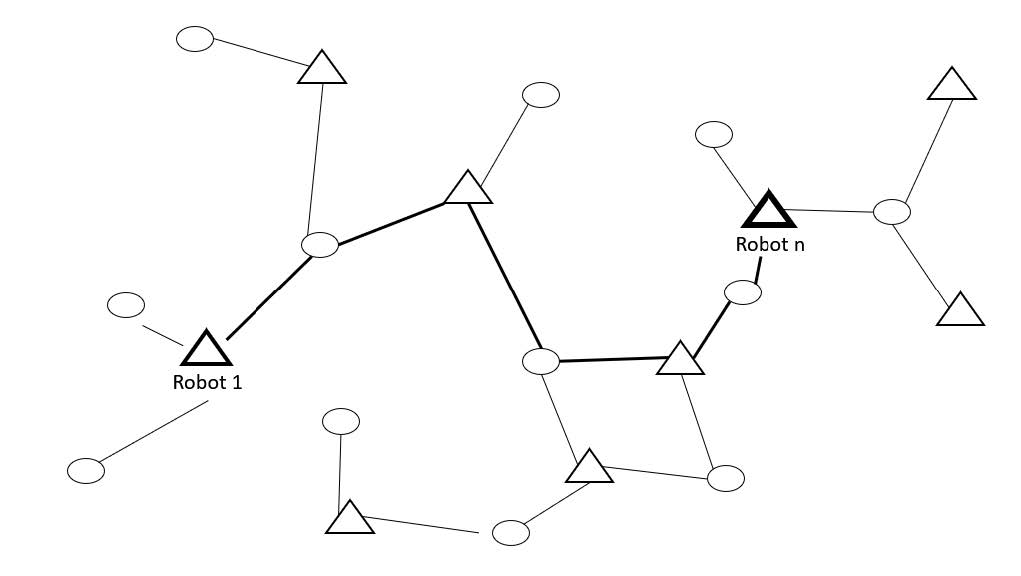}}
\caption{Graph of team of robots (triangles) with visible landmarks (ellipses) for each shown as an edge from robot to landmark. Chain of robots (triangles) with common visual landmarks (ellipses) shown as heavier line.}
\label{fig:pic1}
\end{figure}

Therefore, the research problem of this study is formulated as follows: If a robot is tasked with travelling to a home location, and that location is not in its FOV, then the robot attempts to find another robot in the team that can see the home location and travel towards that robot's location. To stay within the visual homing paradigm, we do not assume that the first robot has access to or can navigate to the spatial coordinates of the second robot. Instead, the two must identify a common visual landmark that can be used as an `intermediate' home location. While this example considers only two robots and one shared visual landmark, the approach generalizes to $n$ robots and $n-1$ landmarks, as shown in the graph in Figure \ref{fig:pic1}, which depicts a team of individual visual homing robots with visible landmarks of which there are several shared landmarks.

\begin{figure*}[htbp]
\centerline{\includegraphics{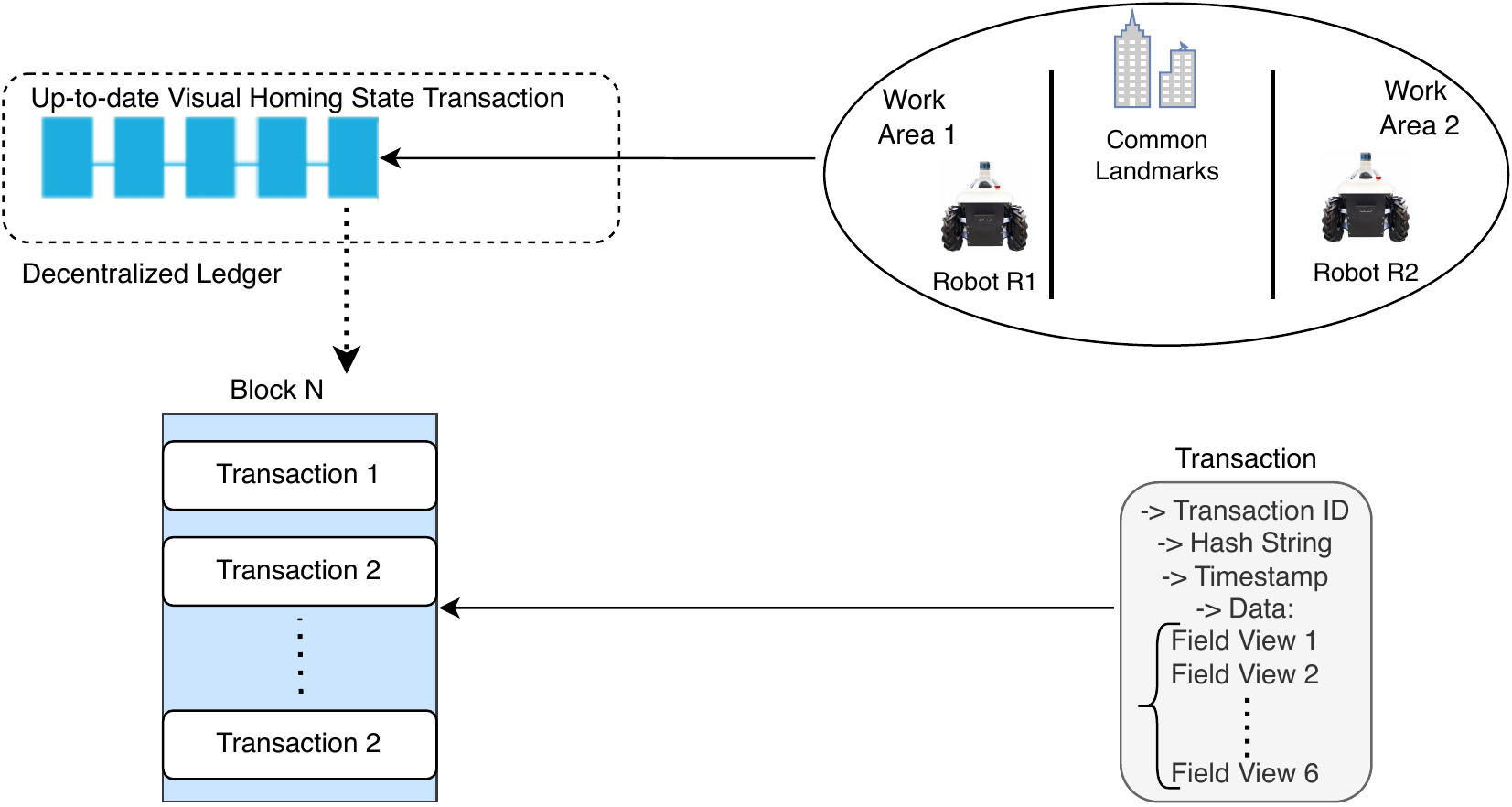}}
\caption{Illustration of the proposed blockchain-enabled visual homing framework.}
\label{fig:pic2}
\end{figure*}

In this work, blockchain technology is adopted to provide decentralized communication among visual homing robots. Such a decentralized communication is trustworthy in terms of data integrity, data confidentiality, and also authentication (i.e., tamper-proof data sharing among robots as they complete their moving task) \cite{rahouti2018bitcoin, ali2019blockchain}. The deployment of blockchain structure in visual homing will guarantee a tamper-proof record of robots' transactions (i.e., individual robot's location record) and rapid retrieval of field views to facilitate the per-robot navigation task \cite{vasylkovskyi2020blockrobot}. One of the most challenging parts of the problem is the analysis of video and identification of common landmarks, and in this paper we focus on building and evaluating that part of the problem.

The remaining of this paper is organized as follows. Section \ref{sec:introduction} provides a background of the research problem and challenges. Section \ref{sec:related} discusses the state-of-art literature related to our work. Section \ref{sec:methodology} presents the methodology and architectural design of the proposed framework, in addition to the research and practical implications related to the framework implementation. Next, Section \ref{sec:evaluation} provides the experimental setups and key evaluation findings demonstrating the efficiency of the proposed solution. Last, Section \ref{sec:conclusion} gives the concluding remarks of this study along with potential future plans.

\section{State of the Art} \label{sec:related}
Visual homing, originally developed as a model of animal behavior \cite{Cartright_1983}, has been extensively studied in robotics, see \cite{nirmal2016homing, Fu_2018, lyons2020evaluation} for a review. Its advantages are that it is lightweight and does not require a metric map data structure or GPS synchronization. The FOV restriction has been addressed by several authors. Steltzer's \cite{Steltzer_2018} Trailmap approach uses a map structure to link several line of sight landmarks together, allowing wide-area navigation. However, in a changing environment, e.g., operating outdoors in all seasons and weathers, such a map would need to be constantly updated and shared. Furthermore, a global map data structure might exceed the memory capacity for a small robot. 

Our approach proposes that robots in a team communicate and identify shared common landmarks, and that these landmarks, which are up to date and path specific, take the place of the intermediate line of sight landmarks in other work. Identifying common visual imagery is as a problem in which deep-learning methods have shown great success. For example, YOLO \cite{Redmon_2018} employs a CNN architecture with 24 convolutional layers followed by two fully connected layers. The image is divided into a fixed-size grid, and when an object is recognized by a grid cell, that grid is responsible for predicting the object class probability and bounding box. This allows YOLO to propose multiple object class matches in a single image pass.

Lyons and Petzinger \cite{Lyons_2020} evaluate several combinations of CNN-based YOLO with SIFT-Based feature recognition to identify common landmarks for two robots in a simulated urban landscape. They that report using Yolo to identify candidate objects and then SIFT to compare candidates yields improved performance over either alone. However, they propose that robots share visual information by point-to-point transmission of imagery. That approach does not scale well to a team and to real deployment. We propose instead using a blockchain approach to decentralize communication among the team. 

Blockchain technology has been integrated into a broad range of modern applications, including, but not limited to, connected and autonomous vehicles (CAVs), Internet of Things (IoT), and robotics \cite{ali2019blockchain}. The deployment of blockchain technology in robotic systems, such as visual homing and navigation, can be highly useful in tackling limitations beyond decentralized/distributed decision-making and security challenges \cite{afanasyev2019blockchain}.

Several notable studies in the literature integrate blockchain technology in robotic-driven applications. For instance, Castello et al. \cite{castello2018blockchain} proposed a blockchain-based framework to improve the security and decision-making in robotic swarm systems, while Fernandes and Alexandre \cite{fernandes2019robotchain} developed blockchain-enabled event management for robotic platforms using Tezos technology. Collective decision-making in robotic systems has further been enhanced through the deployment of blockchain. Namely, in the work by Strobel et al. \cite{strobel2018managing}, the collaborative decision-making problem is addressed in byzantine robots through smart contract-based coordination mechanisms.

Moreover, several research works addressed the data sharing and data monitoring problems in robotic systems. Among these studies, Castello et al. \cite{castello2018robochain} developed RoboChain, a blockchain framework to enhance and secure data-sharing for human-robot interaction. In contrast, Lopes et al. \cite{lopes2019robot} proposed a systematic approach for monitoring the robot workspace using a blockchain-enabled 3D vision mechanism.

According to recent studies \cite{afanasyev2019towards}, even though state-of-the-art algorithms have enabled specialized teams of robots to handle individual-specific behaviors, such as aggregation, flocking, foraging, etc., there is little to no work that integrates blockchain technology into visual homing robotic systems. With its low operational cost, trustworthy functions, provenance, and rigorous access control, blockchain in this work will not only provide enhancements in visual homing systems, but also in new robotic-driven use cases and applications \cite{aditya2021survey}.

To the best of our knowledge, no study has considered the deployment of blockchain technology in visual homing and navigation systems. The proposed framework enables individual robots in a team to efficiently share and identify up-to-date common landmarks in a timely, secure, and trustworthy manner, with low operational cost/overhead.

\section{Methodology and Framework Design} \label{sec:methodology}

This section describes the architectural design and operability of our proposed blockchain-enabled visual homing robotic system.

The proposed solution allows individual robots in a visual homing environment to efficiently share and identify up-to-date common landmarks at a low operational cost and timely manner. Robots are required to create and add a new transaction to the ledger as soon as they complete the move to a new position. A transaction includes the up-to-date visual panorama (i.e., panoramic view) for an individual team robot. The new location (position) is defined as one or more of the following:
\begin{enumerate}
    \item If the robot has traveled over a threshold distance from a prior location.
    \item If the current visual panorama differs by more than a threshold amount from the panorama of the prior location.
    \item If more than a threshold amount of time has elapsed since the prior transaction.
\end{enumerate}

A robot's transaction includes the visual panorama seen from the new robots' location, in addition to a unique hash value. The hash will serve as a unique ID to characterize each transaction in the ledger). The core of the transaction is depicted in Figure \ref{fig:pic2} which consists of the transaction ID, hash string, transaction timestamp, and the core of the transaction. The core of the transaction represents the panoramic view data. Namely, the panoramic view is a sequence (list) of joining images with slightly overlapping fields of view to create the panoramic view for an individual robot. Furthermore, since the ledger consists of a series of blocks that are connected in a chain, each block comprises the core information to be stored, the ``hash" of the information in the block, and the ``hash" of the previous block in the chain. Hashing is done here by converting the string of the transaction's core content into a series of unique numbers and letters. 

\subsection{Motivating Example}
\begin{figure}[htbp]
\centerline{\includegraphics[width=\columnwidth]{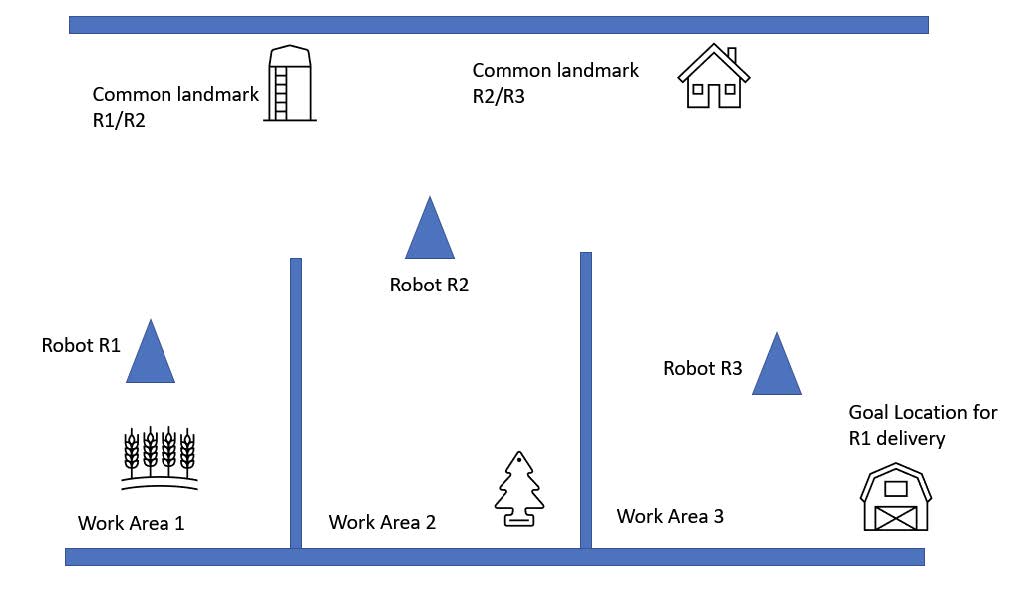}}
\caption{Motivating example.}
\label{fig:pic3}
\end{figure}

A simple motivating example is shown in Figure \ref{fig:pic3}. In this example, $R_1$ needs to deliver its load of grain to a distant barn storage area. $R_1$ knows what the barn looks like, and searches in its visual panorama for a match, but cannot find one. It concludes that the goal location is not in view, and it must utilize network information to travel there.

As discussed, all robots enter their visual panorama information into the blockchain data structure. $R_1$ queries panorama information from the blockchain and identifies that Robot $R_3$ has a visual panorama that includes the goal location for $R_1$ to transport its grain. It then checks if $R_1$ and $R_3$ have any common landmarks. Unfortunately, the geographic landscape (the `walls' in Figure \ref{fig:pic3}) prevent $R_1$ from seeing anything in common with $R_3$. At this point, $R_1$ knows the goal location is in view of $R_3$, but there is still no way for $R_1$ to make its way to the goal.

$R_1$ checks if there is any panorama in the blockchain that has a common landmark with $R_3$. Maybe this information should be precomputed whenever a transaction is added? In this case $R_2$ has a single common landmark with $R_3$. The problem of $R_1$ finding its way to its goal can now be reduced to: can $R_1$ find its way to the common landmark of $R_2$ and $R_3$? If it can, then it could look around, and the goal location should be in view.

$R_1$ next checks if it has a common landmark with $R_2$, and in this case there is one common landmark. $R_1$ now has a potential path to its goal, shown in orange in Figure \ref{fig:pic3}.
\begin{enumerate}
    \item Use visual homing to the common landmark of $R_1$ and $R_2$; this is possible since the common landmark is already in the view of $R_1$ by definition.
    \item When $R_1$ reaches the common landmark, it must look around again and identify the common landmark of $R_2$ and $R_3$.
    \item Use visual homing to common landmark of $R_2$ and $R_3$.
    \item When $R_1$ reaches the common landmark of $R_2$ and $R_3$, look around and identify the goal.
    \item Use visual homing to the goal location.
\end{enumerate}

Informally, this assumes that landmarks are spatially grouped so that if a robot approaches within some distance of one of the landmarks, it will see the other landmarks in that area. The approach distance needs to be chosen so that the  landmark does not occlude some or all the other landmarks.

\subsection{Navigation Use Case}
Here is a use case scenario of how a robot can get to its goal based on the provided use case with three visual homing robots, $R_1$, $R_2$, and $R_3$ in Figure \ref{fig:pic3}:
\begin{enumerate}
    \item $R_1$ will first check which robot ($R_2$ or $R_3$ or both) can see the goal location, and in this case, it is $R_3$ (this task can be easily achieved by looking in the transaction record of the last blockchain block).
    \item $R_1$ checks if it has a common landmark with $R_2$ by looking in the transaction record (in the last block), and in this case there is one common landmark.
    \item $R_1$ next checks if $R_2$ and $R_3$ have a common landmark by looking in the transaction record (in the last block), and in this case there is one common landmark.
    \item Last, $R_1$ should now know the landmark path to get to the goal destination.
\end{enumerate}

\subsection{Panoramic State Update}
As discussed earlier, the blocks are linked through hashing using SHA256. Each block will include the previous block's hash, its own hash, timestamp, and a list of all transactions broadcast by individual robots in the visual homing team. Each time an individual robot in the team changes its location, a new block must be created and appended to the blockchain. The newly created block will include all unchanged transactions (robots remained in the same location) from the previous block, in addition to the new transactions for individual robots that have moved to a new location. Therefore, the blockchain is ensured to maintain the up-to-date panoramic views/states for all individual robots in the last block appended to the ledger.

\section{Evaluation} \label{sec:evaluation}

A similar framework to that of \cite{Lyons_2020} will be used to evaluate our contribution. Pairs of robots will be placed at random locations across 3D simulated urban landscape with spacing 1 to 20m. The quality and amount of common landmarks will be collected as well as the latency and throughput information for the blockchain calculations. The next subsection will describe this testbed in more detail, and the subsequent subsection will present the results.

\subsection{Testbed}

The software for the common landmark recognition testbed is written in the widely-used open-source middleware Robot Operating System (ROS)\footnote{http://www.ros.org}. The 3D simulation engine, Gazebo \footnote{http://gazebosim.org}, has been integrated with ROS to allow simulation testing of robot software. Two Pioneer P3AT robot equipped with cameras are used in conjunction with the modified UCIC python software from \cite{Lyons_2020} for these experiments. The modifications to include the blockchain usage are described below. 

\begin{figure}[htbp]
\centerline{\includegraphics[height=4.5cm, width=8.5cm]{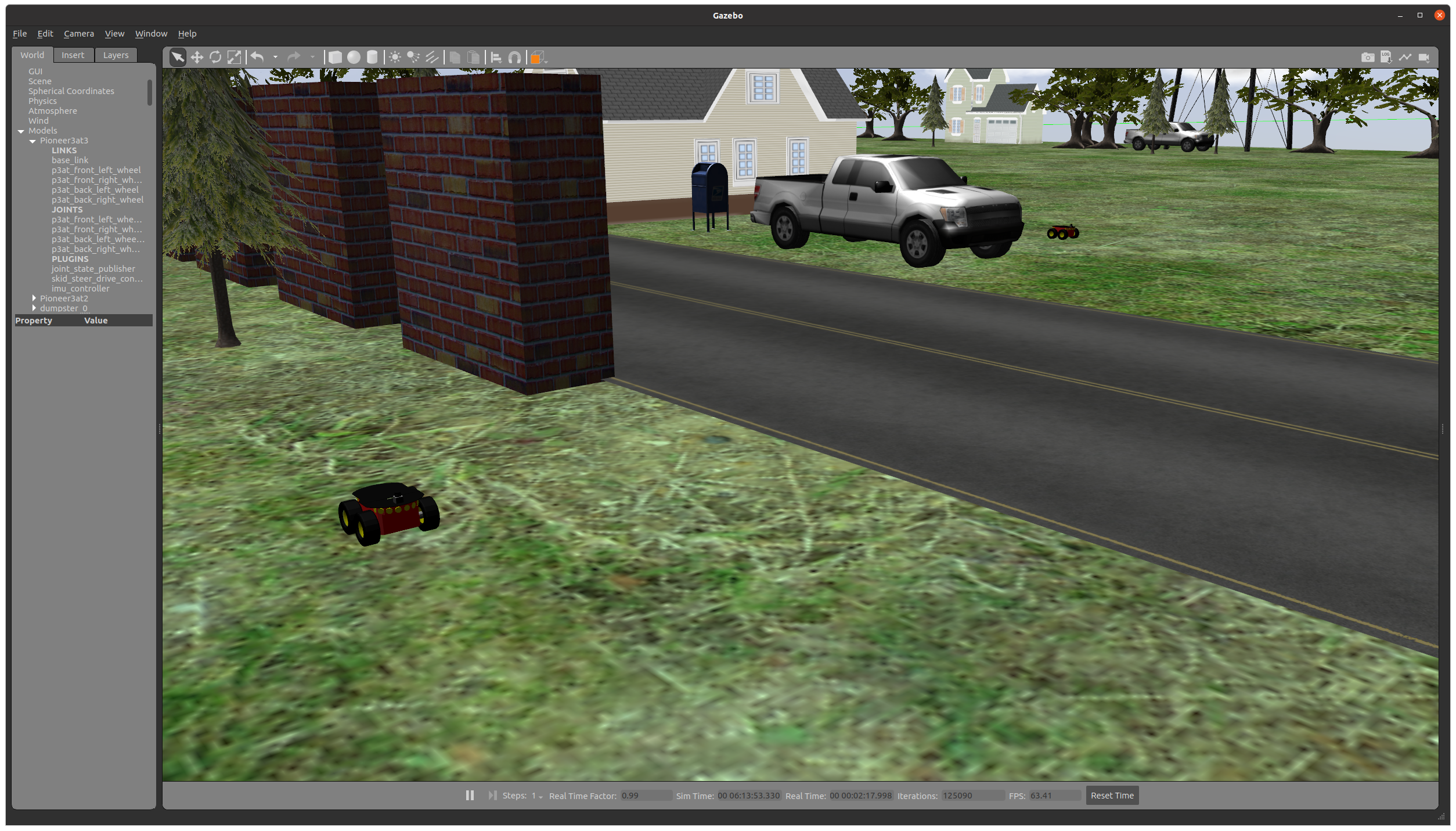}}
\centerline{\includegraphics[height=4.5cm, width=8.5cm]{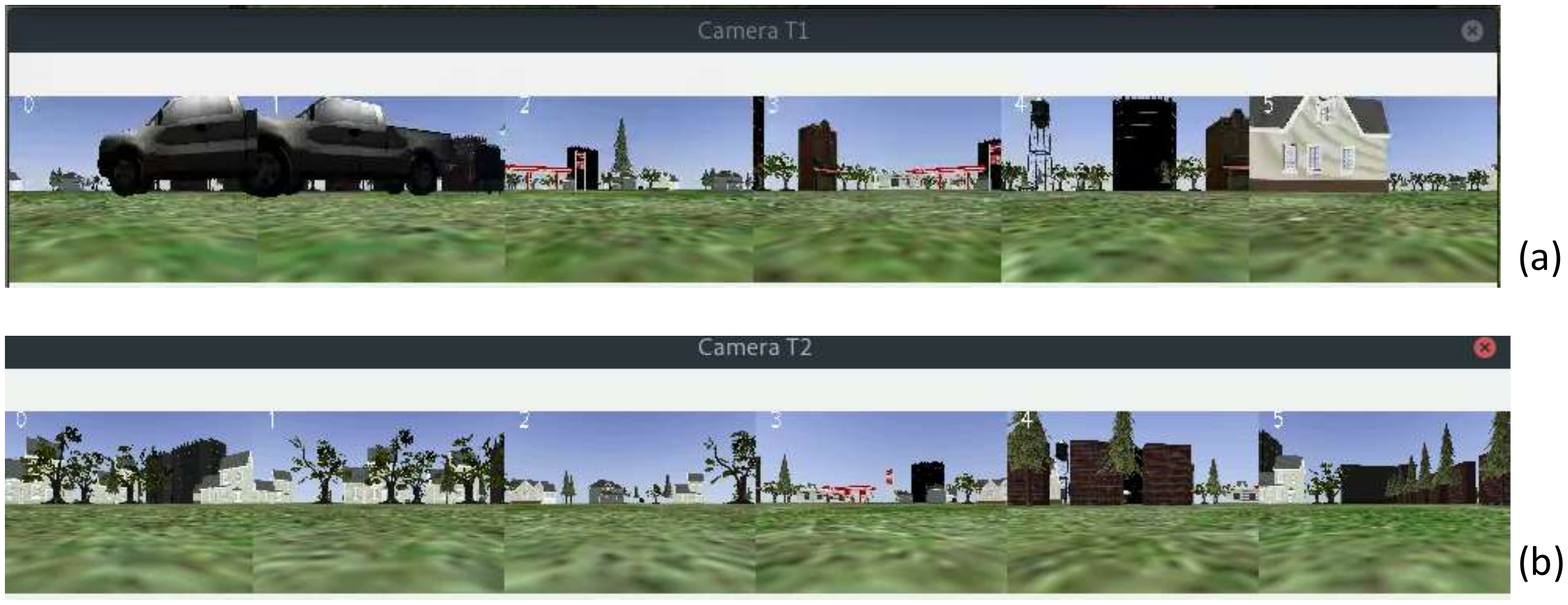}}
\vspace{-1in} 
\caption{Top: Example robot pair positions showing ROS/Gazebo 3D Urban landscape. Bottom: Panoramic views from robot cameras.}
\label{fig:eval1}
\end{figure}

Figure \ref{fig:eval1} shows an example scene from our ROS/Gazebo suburban simulation. The simulation models a $130\times{}180 m^2$ flat, suburban area filled with grass, trees, buildings, vehicles and other objects. The simulation runs on a Digital Storm Aventum with Intel Core-i9 processor and GeForce RTX 3080 GPU.

\subsection{Performance Evaluation}

\begin{figure}[htbp]
\centerline{\includegraphics[height=6.6cm, width=10.5cm]{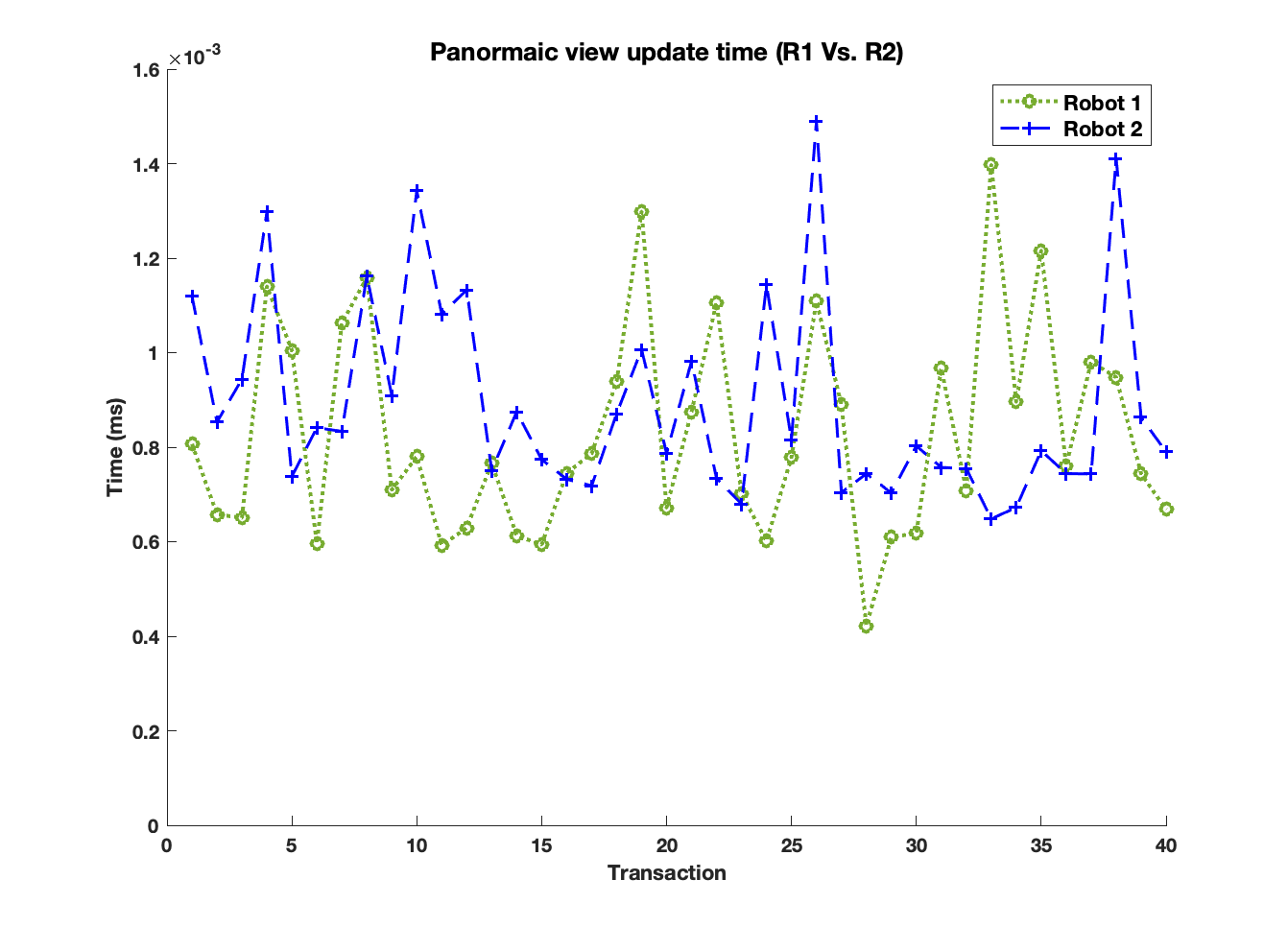}}
\caption{Update time, including transmission time and transaction creation/validation time with 40 positions for robots $R_1$ and $R_2$.}
\label{fig:updatePanView_40}
\end{figure}

\begin{figure}[htbp]
\centerline{\includegraphics[height=7cm, width=10.5cm]{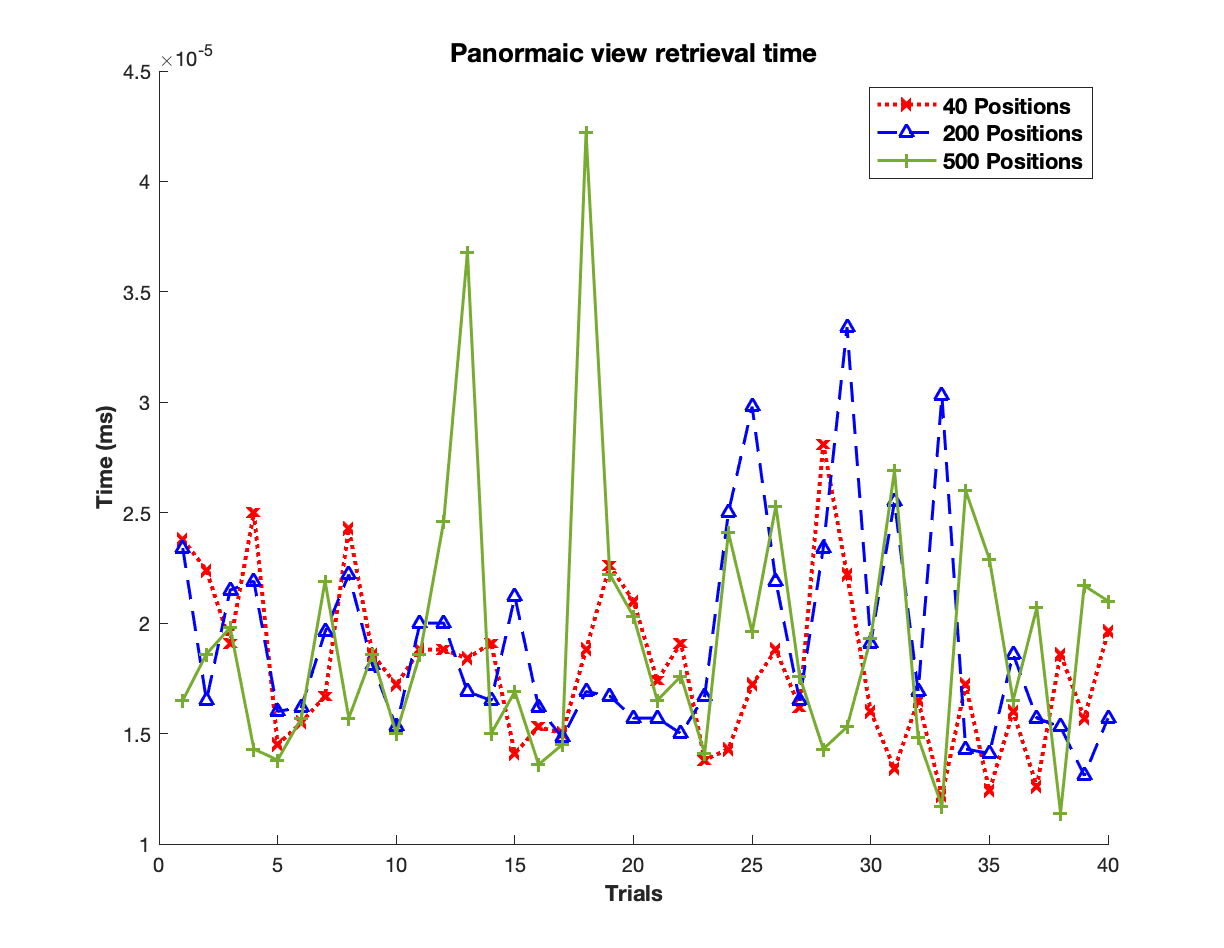}}
\caption{Retrieval time of panoramic views with 40, 200, and 500 positions.}
\label{fig:readPanView_40_200}
\end{figure}

\begin{figure}[htbp]
\centerline{\includegraphics[height=6.7cm, width=10.5cm]{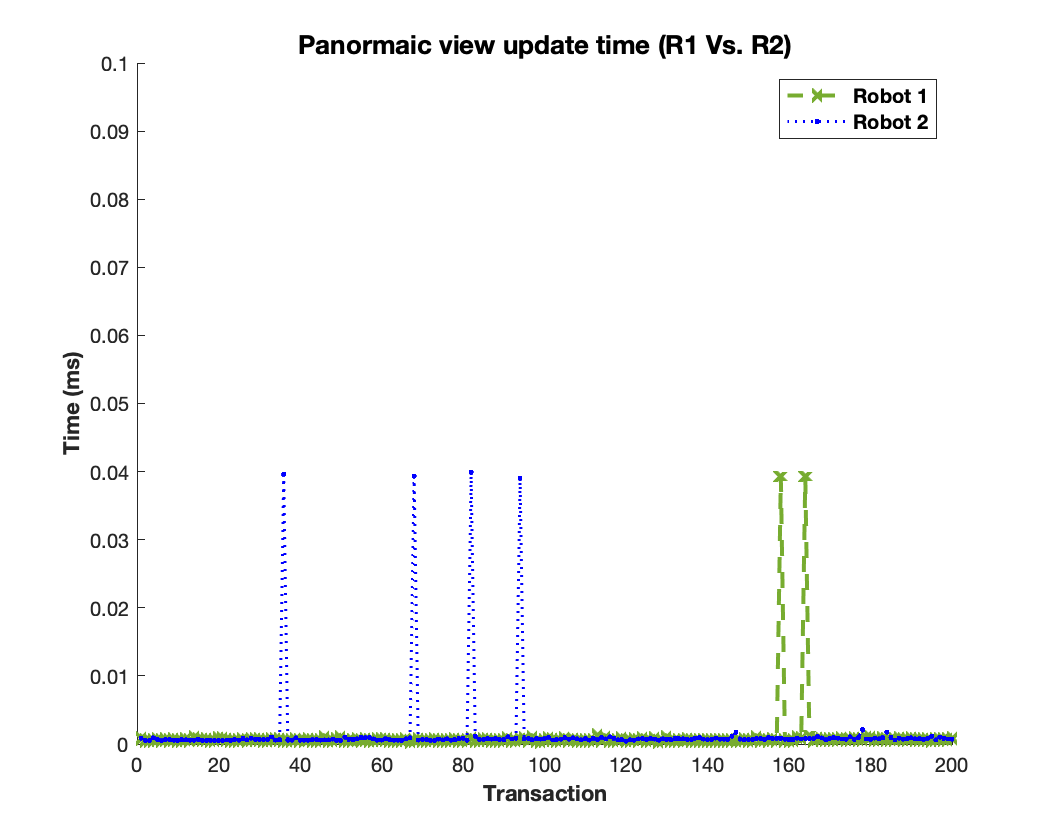}}
\caption{Update time, including transmission time and transaction creation/validation time with 200 positions for robots $R_1$ and $R_2$.}
\label{fig:readPanView_200}
\end{figure}

In order for us to evaluate the efficiency of the proposed solution, we examine the following key metrics: 
\begin{itemize}
    \item Blockchain update time: This represents the time the system takes to update the ledger in accordance with the FoV state changes-- message transmission/broadcast time and transaction validation.
    \item Panoramic view retrieval time: This denotes the time individual team robots take to retrieve a common landmark from the ledger to proceed with the navigation step.
    \item Landmark quality metric: The simulation model database is used to measure how good the common landmark is, as defined in \cite{Lyons_2020}.
\end{itemize}
In this evaluation setup, we examined the proposed framework using varying simulated positions with panoramic views to better demonstrate the system's behavior under a small and large number of transactions (i.e., latency/scalability trade-off).

Figure \ref{fig:updatePanView_40} depicts the blockchain update time in a simulated visual homing environment comprising two team robots, $R_1$ and $R_2$, with 40 different panoramic views. As shown in this figure, the time associated with this operation is bounded by 1.6x$10^{-3}$ ms for both visual homing robots. Similarly, Figure \ref{fig:readPanView_200} shows the blockchain state update time with 200 total panoramic views. The shown time is mostly smaller than 0.01 ms, except in some rare cases where the latency nearly reaches 0.04 ms. These experiments demonstrate the considerably low delay/latency in our solution associated with the blockchain update operation, meaning that when an individual robot moves to a new position, a new transaction containing up-to-date panoramic views can be created and validated in an extremely low time.

Furthermore, Figure \ref{fig:readPanView_40_200} shows the retrieval time in our simulated blockchain-enabled visual homing environment using team robots with 40, 200, and 500 different positions and panoramic views. The plotted time is bounded by 5.5x$10^{-5}$ and mostly smaller than 3x$10^{-5}$ ms, except in some rare cases where it slightly increases for a couple of the 200 and 500 positions trails. Obviously, the latency/communication delay associated with retrieving common landmarks by individual team robots in our blockchain-enabled visual homing system is incomparably small in contrast to state-of-the-art blockchain solutions.

Figure \ref{fig:readPanView_40_200} also shows that the overall number of positions (for the individual team robots) does not remarkably impact the time needed by individual robots to retrieve panoramic views from the blockchain. Last, Table \ref{tab:latency} presents the average delay/communication overhead associated with our blockchain solution over simulated FoVs with varying visual homing robot positions. The table demonstrates the significantly low delay (overall latency) associated with the substantial operations in the visual homing and navigation system, namely, panoramic state update and panoramic state retrieval. It also shows that there has been no loss of average landmark quality as compared to the less sophisticated communication mechanism of \cite{Lyons_2020}.

\subsection{Performance Discussion}

\subsubsection{Latency and Transmission Time Overhead}

It is important to note that distributed solutions for the visual homing and navigation-based robotic systems use classical approaches to broadcast and transmit all panoramic images (of each robot's field of view) to all individual team robots. Therefore, the transmission time complexity associated with broadcasting all panoramas is a $n^2$ graph as the visual homing team increases. In contrast, with a team of $n$ individual robots, our blockchain-enabled solution, the transmission time complexity is incomparably smaller than $n^2$.

The latency in blockchain frameworks may become significant if team robots in a visual homing environment are used in cooperative navigation tasks. Hence, rapid information will be required to orchestrate the movements of the individual robots. Further, collisions may arise when there is a mismatch between the current FoV state and the ledger transaction synchronization. A future extension will be required to improve the latency overhead through affiliation or reputation-based approaches. Team robots belonging to the same FoV region will not need to wait long time frames to accept/process new transactions among themselves.

\subsubsection{Scalability and Throughput}
In our solution, if a large number of robots are deployed within the visual homing and navigation environment, the size of the ledger will be significantly extended (i.e., ledger bloat). Hence, critical parameters such as the block and transaction size and the number of transactions per block, must be optimized in future works \cite{zhou2020solutions}. The blockchain throughput limitation may remarkably impact the FoV state update and retrieval in the case of busy networks with a large number of team robots. A future solution can be the deployment of parallel ledgers with optimized frequency and block size for different FoV information.

\subsubsection{Denial of Service (DoS)}
The DoS issue may arise upon the occurrence of an overwhelming number of team robots interacting with the blockchain, leading to visual homing network disruption. However, the visual homing and navigation environment is generally restricted to small/medium scale, keeping the blockchain infrastructure and network secure against flooding vulnerabilities.

\begin{table}[ht]
\caption{Average latency overhead examination over simulated FoVs with varying visual homing robot positions.}
\label{tab:latency}
\centering
\begin{tabular}{|p{0.7cm}|p{2.1cm}|p{2.1cm}|p{1.8cm}|}
 \hline 
 \# Pos & AVG Ledger Update Time (ms)  & AVG Panoramic View Retrieval Time (ms) & AVG Landmark Quality \\ 
 \hline\hline
 40 & 0.82998 x$10^{-3}$ & 1.79 x$10^{-5}$ & 3.41\\
 \hline
 200 & 0.100515 x$10^{-2}$ & 1.99 x$10^{-5}$  & 3.35\\
 \hline
 500 & 1.09 x$10^{-3}$ & 1.98 x$10^{-5}$ & 3.40\\ 
 \hline
\end{tabular}
\end{table}

\section{Conclusions} \label{sec:conclusion}
Given stored visual information of a target home location, robot navigation back to this location can be accomplished in a light-weight manner using visual homing -- as long that is as the home location is within the FOV, a restriction that limits the generality of this approach. To address this limitation, we consider the robot to be part of a team and we integrate  blockchain technology into the visual homing and navigation system for the team. Based on the decentralized nature of blockchain, the proposed solution enabled team robots to share their visual homing information and synchronously access the stored data of panoramic views to identify common landmarks and establish a navigation path. The evaluation results demonstrated the efficiency of our solution in terms of latency and delay overhead, throughput, and scalability.


\bibliographystyle{IEEEtran}
\bibliography{references.bib}

\begin{thebibliography}{10}
\providecommand{\url}[1]{#1}
\csname url@samestyle\endcsname
\providecommand{\newblock}{\relax}
\providecommand{\bibinfo}[2]{#2}
\providecommand{\BIBentrySTDinterwordspacing}{\spaceskip=0pt\relax}
\providecommand{\BIBentryALTinterwordstretchfactor}{4}
\providecommand{\BIBentryALTinterwordspacing}{\spaceskip=\fontdimen2\font plus
\BIBentryALTinterwordstretchfactor\fontdimen3\font minus
  \fontdimen4\font\relax}
\providecommand{\BIBforeignlanguage}[2]{{%
\expandafter\ifx\csname l@#1\endcsname\relax
\typeout{** WARNING: IEEEtran.bst: No hyphenation pattern has been}%
\typeout{** loaded for the language `#1'. Using the pattern for}%
\typeout{** the default language instead.}%
\else
\language=\csname l@#1\endcsname
\fi
#2}}
\providecommand{\BIBdecl}{\relax}
\BIBdecl

\bibitem{gaussier2000visual}
P.~Gaussier, C.~Joulain, J.-P. Banquet, S.~Lepr{\^e}tre, and A.~Revel, ``The
  visual homing problem: an example of robotics/biology cross fertilization,''
  \emph{Robotics and autonomous systems}, vol.~30, no. 1-2, pp. 155--180, 2000.

\bibitem{castello2018blockchain}
E.~Castell{\'o}~Ferrer, ``The blockchain: a new framework for robotic swarm
  systems,'' in \emph{Proceedings of the future technologies conference}.\hskip
  1em plus 0.5em minus 0.4em\relax Springer, 2018, pp. 1037--1058.

\bibitem{rahouti2018bitcoin}
M.~Rahouti, K.~Xiong, and N.~Ghani, ``Bitcoin concepts, threats, and
  machine-learning security solutions,'' \emph{IEEE Access}, vol.~6, pp.
  67\,189--67\,205, 2018.

\bibitem{ali2019blockchain}
A.~Ali, M.~Rahouti, S.~Latif, S.~Kanhere, J.~Singh, U.~Janjua, A.~N. Mian,
  J.~Qadir, J.~Crowcroft \emph{et~al.}, ``Blockchain and the future of the
  internet: A comprehensive review,'' \emph{arXiv preprint arXiv:1904.00733},
  2019.

\bibitem{vasylkovskyi2020blockrobot}
V.~Vasylkovskyi, S.~Guerreiro, and J.~S. Sequeira, ``Blockrobot: Increasing
  privacy in human robot interaction by using blockchain,'' in \emph{2020 IEEE
  International Conference on Blockchain (Blockchain)}.\hskip 1em plus 0.5em
  minus 0.4em\relax IEEE, 2020, pp. 106--115.

\bibitem{Cartright_1983}
B.~A. Carwright and T.~S. Collet, ``Landmark learning in bees: Experiments and
  models,'' \emph{Journal of Comparative Physiology}, vol. 151, pp. 521--543,
  1983.

\bibitem{nirmal2016homing}
P.~Nirmal and D.~M. Lyons, ``Homing with stereovision,'' \emph{Robotica},
  vol.~34, no.~12, pp. 2741--2758, 2016.

\bibitem{Fu_2018}
F.~Fu and D.~Lyons, ``An approach to robust homing with stereovision,''
  \emph{SPIE Defense \& Security 2017 Conference on Unmanned Systems Technology
  XX}, April 2018.

\bibitem{lyons2020evaluation}
D.~M. Lyons, B.~Barriage, and L.~Del~Signore, ``Evaluation of field of view
  width in stereo-vision-based visual homing,'' \emph{Robotica}, vol.~38,
  no.~5, pp. 787--803, 2020.

\bibitem{Steltzer_2018}
A.~Stelzer, M.~Vayugundla, E.~Mair, M.~Suppa, and W.~Burgard, ``Towards
  efficient and scalable visual homing,'' \emph{Int. J. Robotics Research},
  vol.~37, no. 2-3, 2018.

\bibitem{Redmon_2018}
J.~Redmon and A.~Farhadi, ``Yolov3: An incremental improvement,'' \emph{arXiv},
  2018.

\bibitem{Lyons_2020}
D.~Lyons and N.~Petzinger, ``Visual homing for robot teams: Do you see what i
  see?'' \emph{SPIE 2020 Conference on Unmanned Systems Technology XXIV}, April
  2020.

\bibitem{afanasyev2019blockchain}
I.~Afanasyev, A.~Kolotov, R.~Rezin, K.~Danilov, A.~Kashevnik, and V.~Jotsov,
  ``Blockchain solutions for multi-agent robotic systems: Related work and open
  questions,'' \emph{arXiv preprint arXiv:1903.11041}, 2019.

\bibitem{fernandes2019robotchain}
M.~Fernandes and L.~A. Alexandre, ``Robotchain: Using tezos technology for
  robot event management,'' \emph{Ledger}, 2019.

\bibitem{strobel2018managing}
V.~Strobel, E.~Castell{\'o}~Ferrer, and M.~Dorigo, ``Managing byzantine robots
  via blockchain technology in a swarm robotics collective decision making
  scenario,'' 2018.

\bibitem{castello2018robochain}
E.~Castell{\'o}~Ferrer, O.~O. Rudovic, T.~Hardjono, and A.~S. Pentland,
  ``Robochain: a secure data-sharing framework for human-robot interaction,''
  2018.

\bibitem{lopes2019robot}
V.~Lopes, N.~Pereira, and L.~A. Alexandre, ``Robot workspace monitoring using a
  blockchain-based 3d vision approach,'' in \emph{Proceedings of the IEEE/CVF
  Conference on Computer Vision and Pattern Recognition Workshops}, 2019, pp.
  0--0.

\bibitem{afanasyev2019towards}
I.~Afanasyev, A.~Kolotov, R.~Rezin, K.~Danilov, M.~Mazzara, S.~Chakraborty,
  A.~Kashevnik, A.~Chechulin, A.~Kapitonov, V.~Jotsov \emph{et~al.}, ``Towards
  blockchain-based multi-agent robotic systems: Analysis, classification and
  applications,'' \emph{arXiv preprint arXiv:1907.07433}, 2019.

\bibitem{aditya2021survey}
U.~S. Aditya, R.~Singh, P.~K. Singh, and A.~Kalla, ``A survey on blockchain in
  robotics: Issues, opportunities, challenges and future directions,''
  \emph{Journal of Network and Computer Applications}, vol. 196, p. 103245,
  2021.

\bibitem{zhou2020solutions}
Q.~Zhou, H.~Huang, Z.~Zheng, and J.~Bian, ``Solutions to scalability of
  blockchain: A survey,'' \emph{Ieee Access}, vol.~8, pp. 16\,440--16\,455,
  2020.

\end{thebibliography}

\end{document}